\documentclass[11pt,a4paper]{article}
\usepackage[preprint]{acl}
\usepackage{times}
\usepackage{latexsym}
\usepackage{graphicx}
\usepackage[normalem]{ulem}
\usepackage{booktabs}
\usepackage{pgfplots}
\usepackage{amsmath}

\pgfplotsset{compat=1.18}

\usepackage{url}

\title{Dual-Track CoT: Budget-Aware Stepwise Guidance for Small LMs}

\author{
  Atharva Patil \\
  {\tt adpatil@umass.edu}
  \AND
  Sricharan Ramesh \\
  {\tt sricharanram@umass.edu}
  \AND
  Sagnik Chatterjee \\
  {\tt sagnikchatte@umass.edu}
}
  
\date{}

\begin{document}
\maketitle

\section{Problem statement}

Large Language Models (LLMs) solve many reasoning tasks via chain-of-thought (CoT) prompting, but smaller models (about 7–8B parameters) still struggle with multi-step reasoning under tight compute and token budgets. Existing test-time reasoning methods such as self-consistency (sampling multiple rationales and voting), Tree-of-Thoughts (search over intermediate thoughts), and critique–revise loops improve performance, but often at high token cost and without fine-grained step-level control. This project\footnote{Code and experiments are available at \url{https://github.com/atharvadpatil/DualTrack-COT}.} 
aims to address that gap: \textbf{can Small Language Models (SLMs) reason reliably using the same or fewer tokens?}

This question is both scientific and practical. Scientifically, it probes whether process supervision and simple test-time controls (such as token budgets and rejection of redundant steps) can substitute for model scale or large sampling counts. Practically, many deployments (on-device, low-latency, or cost-constrained settings) cannot afford huge models or dozens of sampled rationales per query. A method that improves SLM reasoning at fixed cost would therefore be directly useful.

\section{What we proposed vs. what we accomplished}

\begin{itemize}
\item Build a Dual CoT pipeline with a Decomposer SLM and an Evaluator SLM working together.
\item Use few-shot prompting to improve accuracy of both models.
\item Modify the GSM8K training dataset to create step-level supervision that matches the framework.
\item Fine-tune the Decomposer on the modified GSM8K dataset.
\item Fine-tune the Evaluator on PRM-800K for step-level scoring.
\item \emph{Planned but not pursued:} scoring on separate criteria such as logical validity, relevance, consistency, complexity, efficiency, and token cost. We dropped this after early experiments showed the evaluator struggled to produce stable multi-criterion scores.
\item \emph{Planned but not kept:} beam search in the final architecture. In practice, beams often collapsed to identical steps generated in each beam through greedy token generation which led to wastage of tokens without improving accuracy.
\item Evaluate the framework on 50 samples from the GSM8K test set.
\item Perform manual error analysis on all test samples.
\item Study the effect of token constraints on accuracy.
\item Introduce a rejection cache to save evaluator tokens on redundant steps and check its impact on accuracy.
\end{itemize}

\section{Related work}

Recent advances in reasoning with large language models (LLMs) have moved beyond linear, single-path generation toward structured or self-corrective paradigms. A prominent branch of work explores search-based reasoning, where models branch and evaluate multiple lines of thought. The \textit{Tree of Thoughts (ToT)} framework \cite{yao2023treeofthoughts}, represents problem solving as a tree of reasoning units, allowing the model to explore, evaluate, and backtrack across paths. This deliberate structure improves performance over standard Chain-of-Thought (CoT) prompting but incurs high computational cost. Expanding and scoring many branches is expensive, limiting its practicality for smaller or resource-constrained models.

In a complementary approach, \textit{Critic-CoT} \cite{zheng2024criticcot} trains LLMs to critique and iteratively refine their own chain-of-thought, evaluating each intermediate step for correctness before revision. However, Critic-CoT mainly operates along a single trajectory, so early mistakes may persist if the model fails to fully correct them; extensions to multi-path variants would likely reintroduce ToT’s computational burden.

Parallel to these efforts, \textit{Self-Consistency} \cite{wang2023selfconsistency} introduces an unsupervised decoding strategy that samples diverse reasoning paths and aggregates final answers by majority vote. It improves accuracy without retraining but suffers from high inference cost and lacks intra-chain correction. The mistakes made early in a chain propagate to the end since errors are never revised mid-way.

More recently, researchers have turned toward process-supervised or verifier-guided reasoning to detect and correct errors at the step level. The \textit{STEPCO} \cite{wu2024stepco} framework integrates a process-supervised verifier (PSV) which scores each reasoning step and triggers local revisions when errors are detected. Although effective, STEPCO generates a full candidate path before verification, meaning tokens might be wasted on later steps following undetected mistakes, and its application in the original work is more focused on numeric or structured tasks.

Two closely related works —  \textit{Math-Shepherd} \cite{wang2024mathshepherd} and \textit{AutoPSV} \cite{lu2024autopsv} — further advance process-supervised reasoning. Math-Shepherd introduces a process reward model (PRM) that assigns rewards to reasoning steps without human labels, using Monte Carlo continuations from each prefix to estimate step quality. This mechanism allows reranking of reasoning traces and reinforcement learning updates via PPO, boosting performance especially on math benchmarks like GSM8K and SVAMP. However, its reliance on large-scale sampling makes it computationally expensive and primarily confined to mathematical reasoning. In contrast, AutoPSV infers process-level supervision via relative changes in verifier confidence. It trains a verifier on final-answer correctness, then uses fluctuations in confidence at each step to infer which steps may be erroneous. AutoPSV thus offers annotation-free process supervision at low cost, though its current form is more oriented toward error detection or reranking rather than stepwise correction.

Despite these advances, most existing systems intervene only after full reasoning chains have been generated. They lack real-time, during-generation guidance and often consume excessive tokens exploring redundant or unpromising reasoning paths. 

\section{Datasets}

\paragraph{GSM8K (Decomposer fine-tuning and evaluation).}

We use \textbf{GSM8K} (Grade School Math 8K)\footnote{\url{https://huggingface.co/datasets/openai/gsm8k}} as our primary benchmark for grade-school math word problems. The dataset contains 8.5K problems paired with worked solutions that include intermediate reasoning and end with an explicit final answer marker. Each example has a \texttt{question} field (natural language problem statement) and an \texttt{answer} field (step-by-step solution in free text). We fine-tune our \textit{Decomposer} on the GSM8K training split and use a small held-out portion as a validation set, keeping the official test split untouched for final evaluation.

\noindent\textbf{Example (GSM8K format).}
\begin{quote}\small
\texttt{"question": "Natalia sold clips to 48 of her friends in April, and then she sold half as many clips in May. How many clips did Natalia sell altogether in April and May?"}\\
\texttt{"answer": "Natalia sold <<48/2 = 24>> clips in May. Natalia sold <<48 + 24 = 72>> clips altogether. \#\#\#\# 72"}
\end{quote}

\paragraph{PRM-800K (Evaluator fine-tuning).}

We fine-tune our \textit{Evaluator} on \textbf{PRM-800K}\footnote{\url{https://huggingface.co/datasets/tasksource/PRM800K}.}, a process reward modeling dataset that provides step-level quality labels for mathematical reasoning. Each completion includes a \texttt{text} field (the candidate step) and a discrete \texttt{rating} label indicating step quality. We train the Evaluator to map a problem and candidate step to this rating, which provides a step-level quality signal. At inference time, we use this capability to accept high quality steps and reject steps that are incorrect, vague, or unhelpful.

\noindent\textbf{Example (PRM-800K completion format).}
\begin{quote}\small
\texttt{"text": "7.8 minutes is the same as 7 minutes and 0.8 minutes.", "rating": 1}\\
\texttt{"text": "Right. 7.8 minutes is the same as 7 minutes and 0.8 minutes.", "rating": 0}
\end{quote}

\paragraph{Why these datasets are challenging.}

GSM8K trains the Decomposer only on correct trajectories: the gold solutions contain successful reasoning steps but no explicit negatives. The model sees what correct next steps look like, but not what wrong steps look like or how to recover from them. As a result, even after fine-tuning, the Decomposer can still make arithmetic mistakes or adopt incorrect decompositions, and there is no explicit training signal for error recovery.

Multi-step error propagation remains a central challenge: an early conceptual or numeric mistake can corrupt the history, so later steps remain locally plausible but lead to the wrong final answer.

For the Evaluator, PRM-800K provides ratings for step quality but no direct supervision for \emph{hints}. At inference time we expect the Evaluator to both score steps and generate short feedback that helps correct low scoring steps. This feedback behavior is only indirectly learned through step ratings, which limits how targeted the hints can be.

\subsection{Data preprocessing}
The raw GSM8K format contains a single worked solution per problem. Our Decomposer is to be trained for iterative next-step prediction, so we convert each example into a set of step-supervised instances. It increases the number of supervised examples per problem, exposes the model to partial histories, and reduces train–test mismatch for stepwise generation.

\paragraph{Step extraction.}
For each example, we parse the \texttt{answer} field into an ordered list of solution steps:

\begin{enumerate}
  \item Remove the trailing final-answer marker (for example the line containing \texttt{\#\#\#\#} and the final numeric answer).
  \item Split the remaining text on line breaks to obtain candidate steps.
  \item Trim whitespace and discard empty lines, preserving inline arithmetic annotations such as \texttt{<<a+b=c>>}, which encode the intended computation.
\end{enumerate}

\paragraph{Next-step supervision.}
Let a problem be \(q\) and its step sequence be \(\{s_1, s_2, \dots, s_T\}\). We generate \(T\) training instances, one per step:
\[
  (q, s_{<t}) \;\mapsto\; s_t, \qquad t = 1,\dots,T,
\]
where \(s_{<t} = (s_1,\dots,s_{t-1})\) denotes the history of completed steps (empty when \(t = 1\)). The input is the original problem plus this history, and the target is the next gold step. This yields dense supervision over the entire solution path and matches how the Decomposer is used at inference time.

\paragraph{Serialized format.}
We serialize each training instance into a rigid textual template with two fields:

\begin{itemize}
  \item \textbf{Problem}: a concatenation of the original question and the current history, for example:
  \texttt{Problem: <question>}\\
  \texttt{Steps completed so far:}\\
  \texttt{STEP 1: <step 1>}\\
  \texttt{STEP 2: <step 2>}\\
  ...
  \item \textbf{Next Step}: the gold next step prefixed with \texttt{STEP:} for intermediate steps and \texttt{FINAL\_ANSWER:} for the final answer step.
\end{itemize}

This format encourages the model to produce exactly one step per query in a consistent style.

\paragraph{Example.}

\begin{quote}\small
    \textbf{Problem}\\
    \texttt{"Problem: Weng earns \$12 an hour for babysitting. Yesterday, she just did 50 minutes of babysitting. How much did she earn?}\\
    \texttt{Steps completed so far:}\\
    \texttt{STEP 1: Weng earns 12/60 = \$<<12/60=0.2>>0.2 per minute."}
    \medskip
    \textbf{Next Step}\\
    \texttt{"STEP: Working 50 minutes, she earned 0.2 x 50 = \$<<0.2*50=10>>10."}
\end{quote}

\section{Baselines}

We compare our approach against four baselines that use the same underlying small instruction tuned Llama model but differ in how they perform reasoning:

\begin{itemize}
    \item \textbf{Plain CoT}: the model receives the full problem and is prompted to output a single answer with chain-of-thought, without any stepwise evaluation or retry loop.
    \item \textbf{Finetuned Decomposer only}: the Decomposer is fine-tuned on GSM8K-style next-step supervision and used to produce full solutions by unrolled step generation, without a separate Evaluator.
    \item \textbf{Dual CoT (few-shot prompting only)}: a two-agent setup where both the Decomposer and Evaluator are used but only with carefully designed prompts and few-shot examples, with no fine-tuning.
    \item \textbf{Finetuned Dual CoT}: the full system with a fine-tuned Decomposer and a fine-tuned Evaluator interacting in a collaborative loop.
\end{itemize}

All baselines are evaluated on the same 50 GSM8K test problems. Hyperparameters for prompting and decoding are kept simple and consistent: low temperature, greedy decoding, and a fixed maximum generation length per query. For fine-tuned models, hyperparameters (learning rate, batch size, number of epochs) are chosen based on validation performance and Colab memory limits.

\section{Approach}

\subsection{High level idea}

Our goal is to improve math word problem solving for small LMs by explicitly separating \emph{step generation} from \emph{step evaluation}. Rather than relying on a single model to both propose and self-check long chains of reasoning, we train two specialized agents on top of the same base model:

\begin{itemize}
    \item a \textbf{Decomposer} that proposes the next reasoning step in a strict textual format;
    \item an \textbf{Evaluator} that scores that step between [0--3] and produces a short piece of feedback that can be reused as a hint.
\end{itemize}

At inference time, these agents interact in a loop. The Decomposer proposes a candidate step conditioned on the problem, the previously accepted steps, and the latest feedback. The Evaluator then scores this step and returns both a scalar quality score and a brief hint. If the score is above a threshold, the step is accepted and appended to the history; if it is below the threshold, the system requests a limited number of revisions, each time passing the hint back to the Decomposer. A global token budget across both agents bounds how much interaction is allowed per problem.

We evaluate two configurations of this loop: a \emph{no-cache} variant that sends every candidate step to the Evaluator, and a \emph{cache-enabled} variant that adds a lightweight rejection cache. In the cache-enabled setting, the solver maintains normalized ``math fingerprints'' of past steps and automatically rejects steps that either (i) repeat an earlier accepted computation or (ii) closely match a step that previously received a very low score. This is intended to save Evaluator tokens on redundant or clearly unhelpful steps while keeping the overall procedure within the same token budget.

\begin{figure}[t]
    \centering
    \includegraphics[width=0.55\textwidth]{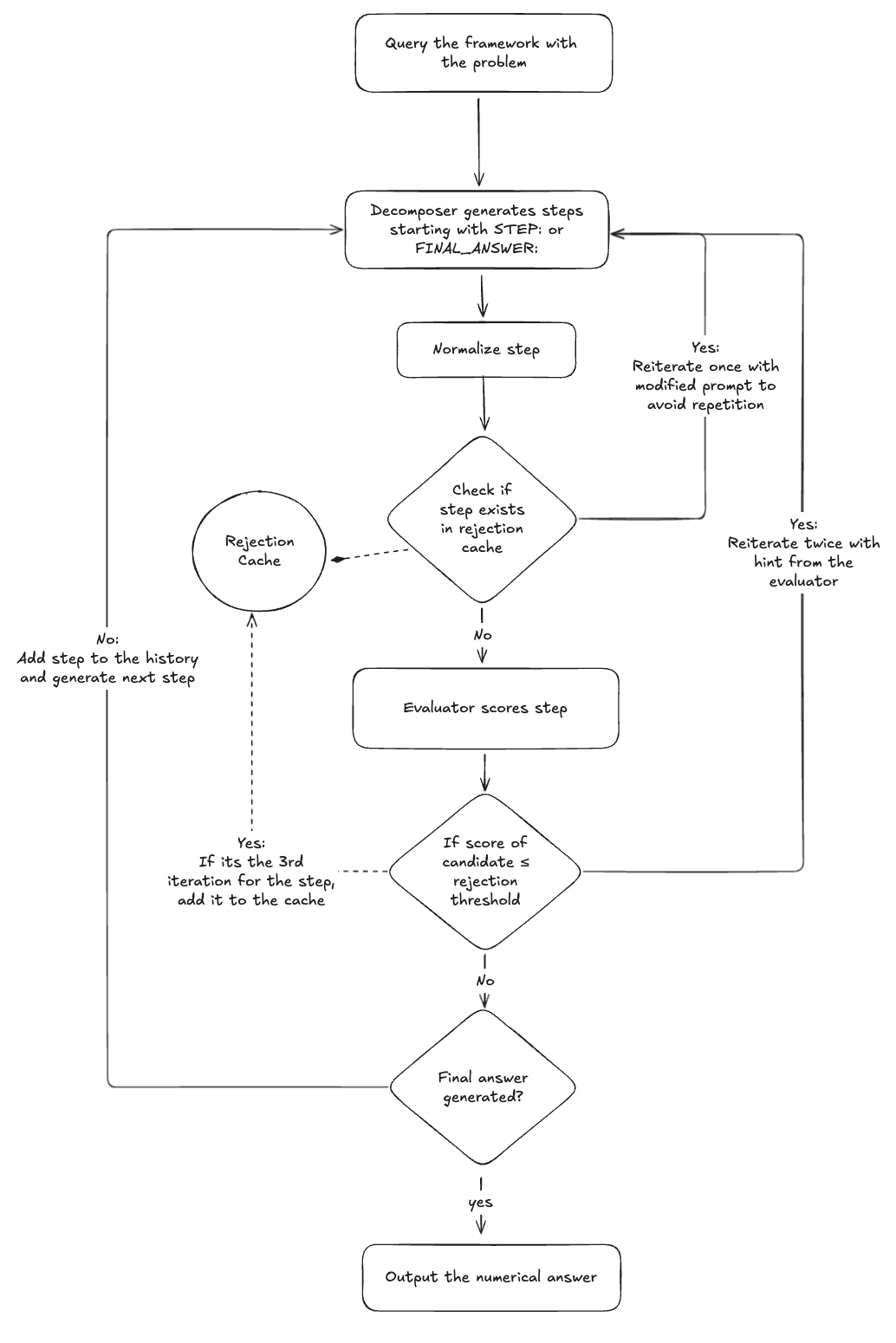}
    \caption{Overall high-level architecture of Dual-Track CoT.}
    \label{fig:architecture}
\end{figure}

\subsection{Collaborative Decomposer–Evaluator loop}

For each problem, the solver maintains an ordered list of accepted intermediate steps.

\paragraph{Step generation.}

The Decomposer receives the problem, the list of accepted steps, and, on retries, a brief hint. It uses a very simple, two-mode output scheme:

\begin{itemize}
    \item On intermediate turns, it returns exactly one line starting with
    \texttt{STEP:}, encoding the next incremental computation or inference.
    \item On the final turn, it returns exactly one line starting with
    \texttt{FINAL\_ANSWER:}; this line contains the numeric solution and is used
    by the solver as a termination signal.
\end{itemize}

This tight format constraint reduces drift and simplifies parsing. Decoding uses low temperature and greedy sampling to keep behavior deterministic and to avoid unnecessary variance.

\paragraph{Repetition guard.}

Before calling the Evaluator, we compute a normalized fingerprint of the current step that keeps only numbers and elementary arithmetic operators and discards most natural language. The same normalization is applied to all previously accepted steps. If the new fingerprint matches a previous one for the same problem, we treat the step as a redundant rephrasing and synthesize a low score and explicit feedback instructing the Decomposer to move to a genuinely new sub-calculation. In this case we do not query the Evaluator, which saves tokens and reduces pointless looping.

\paragraph{Step evaluation.}
If a candidate step is not filtered out by the rejection cache, it is passed to the Evaluator along with the original problem and the current history of accepted steps. The Evaluator returns two outputs: (i) a single integer score \(s \in \{0,1,2,3\}\), and (ii) a short natural language feedback sentence.

The scoring scale is deliberately coarse

\begin{itemize}
    \item \textbf{3 (good step):} mathematically correct, logically follows from the previous steps, and clearly helps move toward the final answer.
    \item \textbf{2 (almost good):} mostly correct and useful, but with a minor imprecision or missing detail that does not invalidate the step.
    \item \textbf{1 (weak / partially correct):} contains some relevant ideas but has an important mistake or omits a key part of the reasoning.
    \item \textbf{0 (bad step):} wrong math, wrong logic, irrelevant or seriously misleading, or effectively the same content as a previous step and therefore not progress.
\end{itemize}

\paragraph{Accept or retry.}

If the overall score exceeds a threshold (\(s > 1\)), the step is accepted and appended to the history. If the accepted step contains a \texttt{FINAL\_ANSWER:} line, the loop terminates and returns that answer. If the score falls below the threshold, the system performs a limited number of retries, each time passing the evaluator feedback as a hint. After a fixed number of retries, the step is accepted even if imperfect, which prevents unbounded oscillation.

\begin{table*}[t]
    \centering
    \begin{tabular}{lccc}
        \toprule
        \textbf{Method} & \textbf{Accuracy} & \textbf{Correct / Total} & \textbf{95\% CI} \\
        \midrule
        Plain CoT             & 78.0\% & 39 / 50 & [64.8\%, 87.2\%] \\
        Dual CoT (few shot, not finetuned) & 24.0\% & 12 / 50 & [14.3\%, 37.4\%] \\
        Finetuned Decomposer               & 68.0\% & 34 / 50 & [54.2\%, 79.2\%] \\
        Finetuned Dual CoT                 & 72.0\% & 36 / 50 & [58.3\%, 82.5\%] \\
        \bottomrule
    \end{tabular}
    \caption{Accuracies and 95\% Wilson confidence intervals on 50 GSM8K-style problems without token constraints.}
    \label{tab:overall-accuracy}
\end{table*}

\paragraph{Token budget.}

A global token budget tracks approximate tokens consumed across both agents. If the budget is exhausted, the solver stops, even if no final answer has been produced. This allows a controlled tradeoff between reasoning depth and cost, and lets us compare cache variants under fixed resource limits.

\paragraph{Rejection cache.}
The cache stores, for each problem, the normalized “math fingerprints” already introduced in the repetition guard: compact representations that collapse different phrasings of the same underlying calculation to a single pattern. For every new candidate step we compute its fingerprint and use the cache in two ways:

\begin{itemize}
    \item \textbf{Redundant accepted steps.} If the fingerprint matches that of a previously \emph{accepted} step, the new step is treated as a redundant restatement of an existing computation and is rejected immediately. The system synthesizes feedback asking the Decomposer to move on to a genuinely new sub-calculation, and the Evaluator is not called.
    \item \textbf{Low-scoring steps.} After evaluation, any step whose raw score falls below a very low threshold (we use a score of 1 or lower on a 0–3 scale) is also added to the cache. This prevents the Decomposer from repeatedly proposing closely similar variants of clearly unhelpful steps and saves further Evaluator calls on those patterns.
\end{itemize}

In the \emph{no-cache} condition, none of these checks are applied: every generated step is sent to the Evaluator, and only the score threshold and retry limit govern whether a step is accepted or revised. In the \emph{cache-enabled} condition, the rejection cache sits in front of the Evaluator, so redundant or obviously bad steps are filtered out early, and the available token budget is spent on more informative evaluations.

\paragraph{Embedding based rejection cache.} We also considered an embedding based variant of the cache that would compare dense sentence embeddings of steps instead of lexical fingerprints. However, this approach risks conflating numerically different steps that are semantically close, for example treating ``10 + 20 = 30'' and ``10 + 20 = 40'' as highly similar in embedding space and rejecting the latter purely because the former was cached as bad.

In a preliminary implementation we used the \url{sentence-transformers/all-MiniLM-L6-v2} SentenceTransformer to build such an embedding based cache. Under the same token budget this variant reached only about 60\% accuracy, compared to 70\% with the simpler lexical cache. To avoid accidentally discarding corrected computations and to keep behavior more predictable, we therefore use only the lexical fingerprint cache in our final experiments and reported results.

\subsection{Fine-tuning setup}
Both agents are obtained by supervised fine-tuning of the same 8B instruction tuned Llama model using the Unsloth framework, on Colab Pro.

\paragraph{Base model and adaptation.}
We load the Unsloth(Quantized) Llama model in 4-bit Quantization using \texttt{bitsandbytes} and \texttt{peft} and apply QLoRA adapters for parameter-efficient fine-tuning. Supervised fine-tuning is implemented with the Hugging Face \texttt{trl SFTTrainer} library.

\paragraph{Decomposer supervision.}
The Decomposer is trained on the custom step-supervised GSM8K data described in Section~4.2, where each example pairs a problem and partial step history with the corresponding gold next step in our strict output format. Here we focus on making this feasible on Colab Pro: we use sequence lengths and batch sizes with gradient accumulation to fit within GPU memory, and train for a small number of epochs until the validation loss stabilizes.

\paragraph{Evaluator supervision.}

The Evaluator is trained on a step rating corpus derived from PRM-style data. Each instance includes a problem, an optional step history, a candidate step, and a scalar rating with a short explanation. We map ratings to a compact numerical range that matches the evaluation prompt and adapt the feedback style to match the desired hint format. The same 4-bit QLoRA setup is used, but the output is structured as multiple score lines plus one feedback line.

\paragraph{Implementation environment.}

All experiments are run on Colab Pro GPUs (T4 and A100). We rely on Unsloth, Hugging Face \texttt{transformers} and \texttt{trl}, \texttt{bitsandbytes}, and \texttt{sentence-transformers}. No external code base implements the collaborative loop, token budget, or rejection caches; these components are implemented within a single notebook.

\subsection{Expected failure modes}

The proposed system inherits some failure modes from standard CoT baselines and introduces new ones. The Decomposer can choose an unhelpful decomposition or make arithmetic mistakes. The Evaluator can mis-score fluent but incorrect steps or fail to notice redundancy. With generous token budgets, the system can loop on repeated restatements instead of making progress. Under tight budgets, the main risk is that early evaluation errors waste the limited retries on a poor trajectory. These behaviors are examined in detail in the error analysis section.

\section{Results}

Our framework is designed to help the Decomposer produce cleaner reasoning: weak steps are down-scored, revised using targeted hints from the Evaluator, and low-quality or repeated steps are filtered out so they do not clutter the history. The example below shows a local correction where a bad step is fixed on retry:

\begin{quote}
\textbf{Step 1}\\
\texttt{STEP: Henry traveled 60 - 45 = 15 miles between his first and second stops.}\\[4pt]

\textbf{Evaluation}\\
\texttt{Raw score: 0.0/3}\\
\texttt{Feedback: The calculation is wrong; he traveled 60 - 20 = 40 miles before his first stop, so he traveled 40 - 15 = 25 miles between his first and second stops.}\\[4pt]

\textbf{Retry (Step 1)}\\
\texttt{STEP: Henry traveled 40 - 15 = 25 miles between his first and second stops.}
\end{quote}

\subsection{Evaluation setup}

We evaluate all models on a held-out set of \(n = 50\) GSM8K-style math word problems. The primary metric is \textbf{answer accuracy}, defined as the proportion of problems for which the final numeric answer exactly matches the gold solution.

We report \textbf{95\% binomial confidence intervals} using the Wilson score interval, which provides better calibrated coverage than the normal approximation for small samples like ours (\(n = 50\)) and proportions away from 0 or 1. For accuracy \(\hat{p}\) and sample size \(n\), with \(z = 1.96\) for a 95\% confidence level, we compute
\[
\tilde{p} = \frac{\hat{p} + \frac{z^2}{2n}}{1 + \frac{z^2}{n}}, \quad
\text{half\_width} =
\frac{
    z \sqrt{ \frac{\hat{p}(1 - \hat{p})}{n} + \frac{z^2}{4n^2} }
}{
    1 + \frac{z^2}{n}
},
\]
and report the interval \([\,\tilde{p} - \text{half\_width},\; \tilde{p} + \text{half\_width}\,]\). Using a fixed \(n = 50\) for all configurations keeps comparisons consistent.

\subsection{Overall performance without token constraints}

Table~\ref{tab:overall-accuracy} summarizes performance when the models are allowed a generous token budget. The non-finetuned Dual CoT model performs poorly at 24\% accuracy, far below the direct-answer baseline at 78\%, and the confidence intervals do not overlap. This suggests that naive, prompted multi-step reasoning can hurt performance when the model is not trained to produce calibrated intermediate steps.

After supervised fine-tuning, the Decomposer-only system reaches 68\% accuracy, closing much of the gap to the direct-answer baseline. The full finetuned Dual CoT configuration reaches 72\% accuracy. The intervals for these two finetuned variants overlap, so the 4 point difference is not statistically decisive at \(n = 50\). Both, however, are clearly stronger than the non-finetuned Dual CoT, and the collaborative setup approaches the direct-answer baseline while exposing explicit intermediate reasoning.

\subsection{Effect of token budget and rejection cache}

\begin{figure*}[t]
    \centering
    \begin{tikzpicture}
        \begin{axis}[
            width=\textwidth,
            height=0.45\textwidth,
            xlabel={Token budget},
            ylabel={Accuracy (\%)},
            xmin=100, xmax=600,
            ymin=0,   ymax=80,
            xtick={100,200,300,400,500,600},
            ytick={0,20,40,60,80},
            grid=both,
            legend pos=south east,
            legend cell align=left,
        ]
            \addplot+[mark=o]
                coordinates {
                    (100,12)
                    (200,46)
                    (300,60)
                    (400,66)
                    (500,66)
                    (600,70)
                };
            \addlegendentry{Without cache};

            \addplot+[mark=triangle]
                coordinates {
                    (100,12)
                    (200,48)
                    (300,60)
                    (400,68)
                    (500,68)
                    (600,70)
                };
            \addlegendentry{With cache};
        \end{axis}
    \end{tikzpicture}
    \caption{Accuracy as a function of token budget for the finetuned Dual CoT system, comparing runs with and without a rejection cache.}
    \label{fig:token-budget-curve}
\end{figure*}
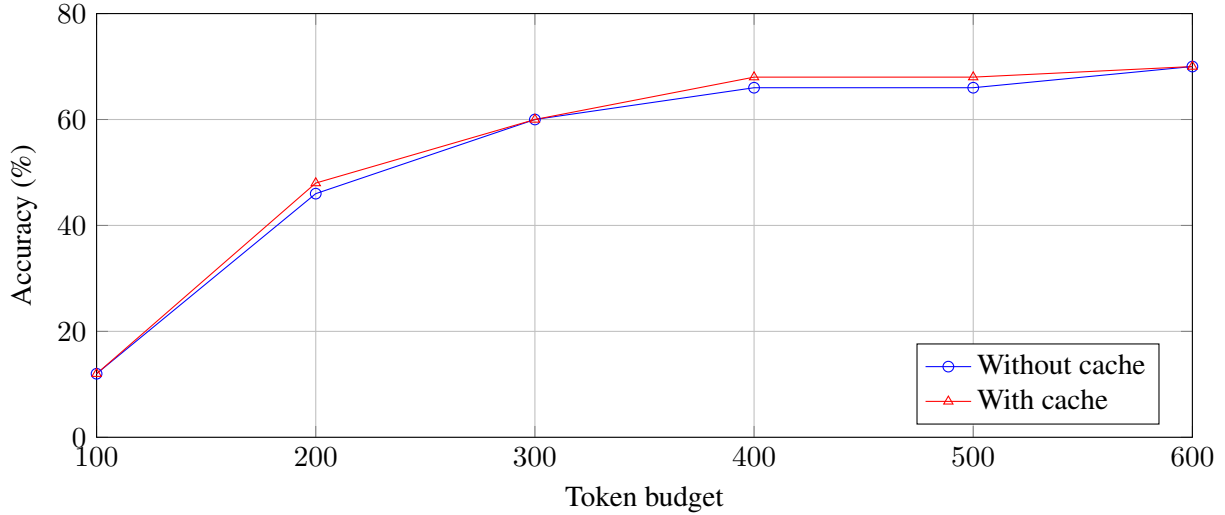

We next vary the global token budget and study the impact of a rejection cache. For each budget, we run the collaborative solver with and without the cache enabled on the same 50 problems. Figure~\ref{fig:token-budget-curve} compares the results.

Accuracy improves steadily with larger budgets. At 100 tokens, both variants achieve only 12\% accuracy. By 300 tokens, accuracy reaches 60\%, and by 600 tokens both variants reach 70\%.

The rejection cache yields small but consistent gains at intermediate budgets: for 200, 400, and 500 tokens the cached variant is 2 points higher. Confidence intervals overlap at all budgets because the sample size per condition is small, but the pattern suggests that filtering out repeated or clearly low-quality steps does not harm accuracy and can help when tokens are scarce. At the largest budget (600 tokens), both variants converge to the same accuracy, which is consistent with the idea that once enough tokens are available the system can often recover even without explicit caching.

\section{Error analysis}

To better understand qualitative behavior, we manually inspected all 50 trajectories produced by the finetuned Dual CoT system in a high-budget setting. 

\paragraph{When the framework succeeds.}
Not all interactions between the Decomposer and Evaluator fail. In many trajectories the loop does exactly what it is designed to do: identify a bad intermediate step, provide targeted feedback, and guide the Decomposer toward a corrected version without restarting the entire solution. For example, in a distance–between–stops problem, the Decomposer initially proposed the incorrect step
\emph{``Henry traveled $60 - 45 = 15$ miles between his first and second stops.''}
The Evaluator assigned a score of $0/3$ and returned specific feedback that Henry first traveled $60 - 20 = 40$ miles before his first stop, so the distance between his first and second stops should be $40 - 15 = 25$ miles. On the next retry, the Decomposer incorporated this hint and revised the step to
\emph{``Henry traveled $40 - 15 = 25$ miles between his first and second stops.''}
This example illustrates that, when feedback is precise and local, the framework can successfully turn an incorrect intermediate step into a correct one, improving the reasoning trajectory rather than merely filtering it.

For each incorrect or low quality solution, we examined the full sequence of Decomposer steps, Evaluator scores, feedback, and the final answer, and assigned one or more labels:

\begin{itemize}
    \item \textbf{Conceptual error (Decomposer)}: the step misinterprets the problem or targets the wrong quantity.
    \item \textbf{Arithmetic error (Decomposer)}: the intended operation is correct, but the numeric computation is wrong.
    \item \textbf{Scoring issue (Evaluator)}: a good step is penalized, a bad step is rewarded, or the feedback is misleading.
    \item \textbf{Looping / no progress}: the system repeats essentially the same content without moving closer to the solution.
    \item \textbf{Ignored hint (Decomposer)}: the Evaluator correctly flags an error and provides a useful hint, but the Decomposer keeps repeating the same flawed pattern instead of incorporating the feedback.
\end{itemize}

We then grouped failures into recurring categories and chose representative examples.

\subsection{Evaluator mis-scoring and weak feedback}

A frequent failure mode is noisy Evaluator supervision: correct steps are discouraged and flawed steps are approved.

\paragraph{Over-penalizing correct steps.}

In several problems the Decomposer produces a valid and relevant step, but the Evaluator treats it as incorrect:

\begin{itemize}
    \item \textbf{Problem 1}: The first step correctly sets up the computation, but the Evaluator marks it wrong and asks for a different logic, pushing the trajectory away from a simple solution.
    \item \textbf{Problem 8}: The initial step is compatible with a reasonable breakdown of the download process, yet the Evaluator claims the logic is wrong and gives an irrelevant hint about load, nudging the Decomposer toward a worse interpretation.
    \item \textbf{Problem 45}: The Decomposer's opening step is correct, but the Evaluator assigns a low score and vague feedback, encouraging unnecessary revisions.
\end{itemize}

Here, correct progress is actively weakened by mis-scoring, which often leads to longer, less stable trajectories.

\paragraph{Accepting conceptually wrong steps.}

More often, the Evaluator focuses on local arithmetic and wording and fails to detect that the step answers the wrong question:

\begin{itemize}
    \item \textbf{Problem 2}: The Decomposer misreads \(150\%\) as \(0.15\) instead of \(1.5\). The Evaluator does not correct this and scores the step as acceptable.
    \item \textbf{Problem 16}: After computing profits for jewelry and gadgets, the Decomposer subtracts them to answer ``how much more profit'' one option yields, even though the question asks for total profit. The Evaluator treats this as a correct comparison.
    \item \textbf{Problem 22}: The Decomposer hallucinates specific calendar years and reasons with them, even though the prompt only provides ages. The Evaluator accepts the step because the internal arithmetic within the hallucinated timeline is consistent.
    \item \textbf{Problem 39}: The model builds an incorrect notion of average speed and uses it throughout, while the Evaluator repeatedly approves steps that contain fluent but conceptually wrong reasoning.
    \item \textbf{Problem 46}: The Decomposer computes \(5 \times \tfrac{2}{5} = 4\) and misinterprets ``\(\tfrac{2}{5}\) times more'' as a stand alone multiple. The Evaluator overlooks both issues and approves the step.
\end{itemize}

In these cases, the Evaluator's focus on local correctness is not sufficient to guard against global conceptual drift.

\paragraph{Failure to recognize loops and lack of progress.}

In some trajectories the Evaluator never signals that the system is stuck:

\begin{itemize}
    \item \textbf{Problem 24}: The Decomposer repeats the same sentence for multiple steps. The Evaluator keeps assigning high scores with positive feedback, so no progress is made despite high token usage.
    \item \textbf{Problem 36}: The correct answer is reached, but the system continues to re-derive or paraphrase the same key step instead of emitting a concise final answer.
\end{itemize}

These loops are more prominent when the token budget is large, since there is no external pressure to stop.

\subsection{Conceptual misunderstandings in decomposition}

The Decomposer also exhibits recurring conceptual errors, especially in problems that require tracking quantities over time.

\paragraph{Generating the wrong quantity.}

\begin{itemize}
    \item \textbf{Problem 13}: The model correctly finds the break-even year when cumulative profit equals initial cost, but returns this year as the answer instead of the first year of actual profit.
    \item \textbf{Problem 16}: After computing each profit component correctly, the model answers a comparison question (difference in profit) rather than the requested total profit.
\end{itemize}

In both examples, local computations are reasonable but the final quantity is misaligned with the question.

\paragraph{Misinterpreting logical structure.}

\begin{itemize}
    \item \textbf{Problem 8}: The Decomposer treats ``load'' as average rate and divides file size by elapsed time, instead of summing time segments before and after a pause in the download.
    \item \textbf{Problem 9}: When the person turns back home, the model adds the return distance to the outward trip instead of subtracting it, so the distance from home increases instead of decreasing.
    \item \textbf{Problem 17}: Distances traveled by two trains are combined and then partially subtracted to represent ``traveling together,'' which does not match the spatial setup.
    \item \textbf{Problem 38}: The Decomposer starts correctly by computing remaining money, then misuses an unrelated change amount when trying to infer the number of Lego sets, leading to a wrong count.
    \item \textbf{Problem 47}: The model ignores conservation of items and uses a formula of the form \(start + used - remaining\), instead of \(start + bought = used + remaining\).
\end{itemize}

These failures cluster in narrative problems involving motion, rates, inventory, or money flows. The model often loses track of a conserved quantity or misapplies a relationship between stages.

\subsection{Arithmetic errors that survive evaluation}

Some failures are simple numeric mistakes that neither agent corrects.

\begin{itemize}
    \item \textbf{Problem 2}: The misinterpretation of \(150\%\) as \(0.15\) propagates through the entire solution, and the Evaluator does not flag the inconsistency.
    \item \textbf{Problem 44}: The Decomposer combines multiple intermediate calculations into one line and miscomputes part of the expression. The Evaluator assigns a low score but focuses criticism on the wrong aspect, so the error remains.
    \item \textbf{Problem 46}: The product \(5 \times \tfrac{2}{5}\) is miscalculated as 4 instead of 2, and this mistaken value is treated as correct by the Evaluator.
\end{itemize}

These errors are less frequent than conceptual mistakes but show that explicit numeric supervision still leaves gaps when math is embedded in free text.

\subsection{Correct evaluator signal, ignored by the Decomposer}

A distinct pattern emerges in cases where the Evaluator does its job reasonably well, but the Decomposer fails to make use of the feedback. Here the bottleneck is not detection of the error, but the system's ability to repair it.

\begin{itemize}
    \item \textbf{Problem 47}: The Evaluator correctly points out that the conservation equation is missing the 23 post-it notes and asks the model to account for them. The Decomposer nevertheless keeps reusing the same incorrect structure \(start + used - remaining\) for several retries, effectively ignoring the hint. After enough retries, the Evaluator relaxes its score and accepts a step that still encodes the wrong logic.
    \item \textbf{Milder cases}: In a few other problems, the Evaluator downgrades a step and highlights a specific flaw (for example, missing a time segment or misreading a ``times more'' phrase), but the next Decomposer step only slightly rephrases the previous attempt instead of fixing the identified issue.
\end{itemize}

In these trajectories, the Evaluator provides a correct local gradient, but the Decomposer does not follow it. This suggests that simply feeding back natural-language hints is not always enough to induce robust self-correction, especially when the underlying plan needs to change rather than a single arithmetic detail.

\subsection{Comparison to baselines}

Relative to the baselines, our system fails in systematically different ways:

\begin{itemize}
    \item The \textbf{direct-answer baseline} mainly fails on problems that genuinely require multi-step reasoning. It tends to make a single early algebraic slip or misread a phrase like “times more,” and never revises the mistake.
    \item The \textbf{non-finetuned Dual CoT} produces noisy, off-topic intermediate steps, lacks stable decompositions, and rarely produces the \texttt{FINAL\_ANSWER:} keyword despite reaching the final answer, making it consistent with its much lower overall accuracy.
    \item The \textbf{finetuned Dual CoT} system typically gets local computations right and produces clean-looking steps, but is vulnerable to conceptual drift, evaluator mis-scoring, and looping. In many failed cases the correct answer is one or two conceptual fixes away, yet the interaction fails to recover.
\end{itemize}

\section{Contributions of group members}

\begin{itemize}
    \item \textbf{Atharva Patil}
    \begin{itemize}
        \item Project ideation and conceptualization.
        \item Design and implementation of the Dual CoT framework.
        \item Experimental Design and Prompt Engineering (evaluation of diverse prompts, few-shot configurations, and scoring schemes)
        \item Qualitative error analysis of trajectories generated by the fine-tuned Dual CoT system for Problems 1--15.
    \end{itemize}

    \item \textbf{Sricharan Ramesh}
    \begin{itemize}
        \item Baseline model analysis.
        \item Design and implementation of the rejection cache.
        \item Token budget management with variants and optimization.
        \item Qualitative error analysis of trajectories generated by the fine-tuned Dual CoT system for Problems 31--45.
    \end{itemize}

    \item \textbf{Sagnik Chatterjee}
\begin{itemize}
    \item Fine-tuning and optimization of language models for structured reasoning.
    \item Construction and curation of training and evaluation datasets.
    \item Systematic benchmarking of the fine-tuned Dual CoT framework against the base Llama model.
    \item Qualitative error analysis of reasoning trajectories produced by the fine-tuned Dual CoT system on Problems 16--30 and 46--50.
\end{itemize}

\end{itemize}

\section{Conclusion}

In this project, we investigate from first principles how a small language model can be transformed into a structured mathematical reasoner. We begin with a direct-answer baseline using the quantized Llama 3.1 8B model, and then introduce a supervised Decomposer module that generates explicit intermediate reasoning steps, evaluating its impact on performance. We subsequently integrate an Evaluator, initially prompted using few-shot examples and later fine-tuned on step-level quality annotations. By inspecting complete reasoning trajectories, we analyze both the conditions under which the combined system improves performance and the failure modes that limit its effectiveness. When accuracy gains plateau despite extensive prompt engineering, we study token-budget constraints and demonstrate that a simple rejection cache can yield modest improvements at moderate budgets by filtering repeated or clearly low-quality intermediate steps.

\paragraph{Future work} We could strengthen the Evaluator by enabling it to produce structured, corrective feedback rather than only scalar quality scores. In the current formulation, the Evaluator is trained solely on step-level ratings and does not receive supervision on how erroneous steps should be revised. A natural extension is to equip the Evaluator with a hint-generation interface, in which it outputs both a score and a concise, targeted explanation of what is incorrect, incomplete, or conceptually misaligned in a given step. Training the Evaluator on hint-style or correction-focused datasets could make its feedback more actionable and better aligned with human tutoring signals.

\begin{quote}
\textbf{Input:} \\
\texttt{Problem: <question>}\\
\texttt{Steps completed so far:}\\
\texttt{STEP 1: <step 1>}\\
\texttt{STEP 2: <step 2>}\\
... \\

\textbf{Output:} \\
\texttt{Score: <0--3>} \\
\texttt{Feedback: <short natural language hint>} \\
\end{quote}

Building on this, the Decomposer could be fine-tuned directly on feedback for revising intermediate steps. Instead of training only on correct gold steps, we can construct a dataset where the input includes the problem context, a prior low-quality step, its score, and a hint from an Evaluator, and the target is a revised, higher-quality step. This would turn feedback from the Evaluator into a concrete training signal for local correction, rather than just a test-time heuristic.

\begin{quote}
\textbf{Input:} \\
\texttt{Problem: <question>}\\
\texttt{Steps completed so far:}\\
\texttt{STEP 1: <step 1>}\\
\texttt{STEP 2: <step 2>}\\
\texttt{SCORE: <score for step 2>}\\
\texttt{HINT: <hint to improve step 2>}\\

\textbf{Output:} \\
\texttt{STEP: <higher quality step>} \\
\end{quote}

Finally, we plan to explore the integration of Tiny Recursive Models (TRMs), which introduce explicit recursive refinement through small, specialized submodels that iteratively improve intermediate reasoning states \cite{trm}. Such recursive architectures align naturally with our step-based framework and may help mitigate conceptual drift by enabling structured, multi-pass correction rather than one-shot evaluation.

Overall, our findings suggest that collaborative, step-based reasoning frameworks are a promising direction for enhancing the capabilities of small language models. However, achieving robust and scalable improvements will likely require progress-aware evaluation mechanisms and stronger supervision for detecting global reasoning errors, not merely local inconsistencies..

\section{AI Disclosure}
\begin{itemize}
    \item Did you use any AI assistance to complete this proposal? If so, please also specify what AI you used.
    \begin{itemize}
        \item GPT 5 by OpenAI
        \item Gemini by Google
    \end{itemize}
\end{itemize}

\noindent\textit{If you answered yes to the above question, please complete the following as well:}

\begin{itemize}
    \item  If you used a large language model to assist you, please paste *all* of the prompts that you used below. Add a separate bullet for each prompt, and specify which part of the proposal is associated with which prompt.
    \begin{itemize}
        \item \textbf{Approach}: Adding beams in step generation provides the same text given problem and previous steps, is there a way to generate a variety of steps without introducing hallucination?
        \item \textbf{Approach}: How to perform peft with quantized model
        \item \textbf{Approach}: How to improve arithmetic reasoning of the Llama 3.1 8B through prompting?
        \item \textbf{Approach}: Can we normalize the step by removing words to avoid cache miss?
        \item \textbf{Result}: Explain Wilson Confidence Intervals
    \end{itemize}
    \item \textbf{Free response:} For each section or paragraph for which you used assistance, describe your overall experience with the AI. How helpful was it? Did it just directly give you a good output, or did you have to edit it? Was its output ever obviously wrong or irrelevant? Did you use it to generate new text, check your own ideas, or rewrite text?
    \begin{itemize}
        \item The LLMs provided good and helpful inputs when it came to improving and optimizing code, but it was not very insightful with the architecture and improvement of prompts of the models. It was also helpful in analyzing the baselines quantitatively and qualitatively.
    \end{itemize}
\end{itemize}

\footnotesize
\bibliography{yourbib}

\end{document}